# Enhancing Few-Shot Learning with Integrated Data and GAN Model Approaches

Yinqiu Feng
Columbia University
New York, USA

Aoran Shen
University of Michigan
Ann Arbor, USA

Jiacheng Hu
Tulane University
New Orleans, USA

Yingbin Liang
Northeastern University
Seattle, USA

Shiru Wang
Dartmouth College
Hanover, USA

Junliang Du*
Shanghai Jiao Tong University
Shanghai, China

*Abstract*— **This paper presents an innovative approach to enhancing few-shot learning by integrating data augmentation with model fine-tuning in a framework designed to tackle the challenges posed by small-sample data. Recognizing the critical limitations of traditional machine learning models that require large datasets—especially in fields such as drug discovery, target recognition, and malicious traffic detection—this study proposes a novel strategy that leverages Generative Adversarial Networks (GANs) and advanced optimization techniques to improve model performance with limited data. Specifically, the paper addresses the noise and bias issues introduced by data augmentation methods, contrasting them with model-based approaches, such as fine-tuning and metric learning, which rely heavily on related datasets. By combining Markov Chain Monte Carlo (MCMC) sampling and discriminative model ensemble strategies within a GAN framework, the proposed model adjusts generative and discriminative distributions to simulate a broader range of relevant data. Furthermore, it employs MHLoss and a reparameterized GAN ensemble to enhance stability and accelerate convergence, ultimately leading to improved classification performance on small-sample images and structured datasets. Results confirm that the MhERGAN algorithm developed in this research is highly effective for few-shot learning, offering a practical solution that bridges data scarcity with high-performing model adaptability and generalization.**

*Keywords; Few-Shot Learning, Data Augmentation, Model Fine-Tuning, Meta-Learning, Small-Sample Data Analysis.*

## I. Introduction

Currently, the effectiveness of most machine learning models relies on a large amount of data. However, in practical fields such as drug discovery [1], medical health records [2-3], and malicious traffic detection [4], it is difficult to obtain a large amount of effective data due to the confidentiality, scarcity, and high cost of data acquisition inherent to these domains. Constructing machine learning models with good performance using a small amount of data facilitates the widespread application of models in various practical scenarios. How to extract more accurate information from small-sample data and ensure the generalization performance of models is one of the current research hotspots.

Few-shot learning can be studied from three perspectives: data, models, and optimization algorithms. Data-based few-shot learning methods aim to solve the small-sample problem by enhancing the dataset, which can be divided into sample-level methods and feature-level methods [5]. Sample-level methods mainly expand the sample quantity; feature-level methods mainly reduce the required sample size through feature selection and optimization [6]. Model-based few-shot learning methods address the small-sample problem from the perspective of model design, which can be divided into model fine-tuning and metric learning methods [7]. Model fine-tuning methods leverage knowledge learned in related domains to improve the performance of few-shot learning; metric learning [8] methods aim to learn suitable metrics for small-sample data to uncover more valuable information [9]. Optimization algorithm-based few-shot learning methods improve the probability of finding the optimal hypothesis by modifying the optimal hypothesis search approach.

At present, the widely recognized optimization algorithm suitable for few-shot learning is meta-learning. The core of meta-learning is to modify the optimization algorithm, training a meta-learner with multiple tasks, which aims to minimize the sum of losses across all tasks. The goal is to endow the model with the ability to learn and quickly learn new tasks.

For few-shot learning, data-based methods are the most intuitive but may introduce noise or bias, while model-based methods and optimization algorithm-based methods mostly require the assistance of related datasets.

Combining data-based methods with methods from the other two perspectives can simultaneously address the issues of bias introduction and the acquisition of related datasets. Therefore, this paper takes small-sample data as the research object, focusing on the bias introduction issue of data augmentation methods and the related data acquisition problem of model fine-tuning methods. It comprehensively considers both data and model perspectives to explore a framework and algorithm that can appropriately combine the two types of methods, aiming to extract more accurate information from limited samples and obtain a model with good performance.

## II. BACKGROUND

Generative Adversarial Networks (GANs) [10] learn data distributions and generate new data using maximum likelihood principles and adversarial learning from game theory.

### A. GAN

As shown in Figure 1, a GAN consists of two parts: a generator G and a discriminator D. The generator aims to learn the data distribution [11], while the discriminator aims to correctly distinguish between real and generated samples [12]. Through adversarial training between discriminator D and generator G, the model learns the dataset distribution and generates new samples.

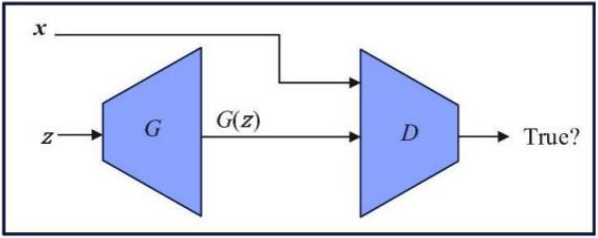

Figure 1. GAN

The objective function is shown in equation (1), where $D(x)$ represents the probability that discriminator classifies real sample $x$ as real, and $D(G(z))$ represents the probability that discriminator classifies generated sample $G(z)$ as real. From equation (1), we can see that discriminator D aims to maximize the distance between generated and original data distributions, while generator G aims to minimize the maximum distance between generated and original data distributions.

$$\begin{aligned}
&\min G \max DV(G,D) \\
&= Ex \sim px[\log D(\mathbf{x})] \\
&\quad - E_{\mathbf{z} \sim p_z}[\log(1 - DG\mathbf{z})] \, E_{\mathbf{x} \sim p_x}[\log D\mathbf{x}] \\
&= \int p_\mathbf{x} \mathbf{x} \log D\mathbf{x} d\mathbf{x}
\end{aligned} \quad (1)$$

### B. WGAN

The original GAN had several issues. WGAN (Wasserstein Generative Adversarial Network) [13] studied GAN theoretically and provided solutions to problems such as inadequate distance metrics for model training [14], slow convergence, and mode collapse. WGAN's modifications include three aspects: removing the sigmoid layer [15], replacing JS distance with Wasserstein distance, and adding weight smoothness constraints [16]. WGAN's objective function is shown in equation (2), where $|D|_L \leq K$ indicates that discriminator D is a K-Lipschitz function.

$$\begin{aligned}
&\min G \max D, |D|L \leq KV(G,D) \\
&= Ex \sim p_x[D(\mathbf{x})] - E_{\mathbf{z} \sim p_z}[(1 - D(G(\mathbf{z})))]
\end{aligned} \quad (2)$$

## III. METHOD

For few-shot data, the distributions learned by both the generator and discriminator in classical generative adversarial networks exhibit biases [17]. To address this issue, this paper introduces the reparameterization GAN method into few-shot learning, employing MCMC sampling to correct the distribution learned by the generator, while using ensemble methods to constrain discriminator learning and correct its learned distribution.

### A. Few-shot Learning Architecture Based on GAN Ensemble and MHLoss Fine-tuning

This paper extends the DAMFT_FSL framework by introducing a reparameterization GAN ensemble method into the data augmentation module to correct the distributions learned by both generator and discriminator models, thereby obtaining more relevant datasets. It also incorporates increased iteration rounds and MHLoss [18] strategy into the model fine-tuning module to enhance fine-tuning stability and achieve better classification performance.

Specifically, an ensemble strategy is added to the discriminator of the generative adversarial network to obtain a well-trained generator and a more accurate discriminator on few-shot datasets [19]. MCMC sampling is then used to correct the distribution learned by the generator [20] to obtain the relevant dataset. Based on the classifier pre-trained using the relevant dataset, the classification model is fine-tuned through increased iteration rounds and MHLoss strategy. The DAMFT_FSL2 architecture based on reparameterization GAN ensemble and MHLoss fine-tuning is shown in Figure 2.

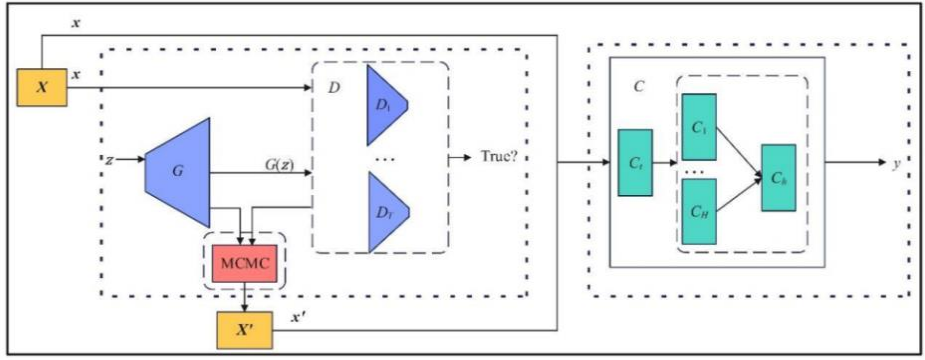

Figure 2. DAMFT_FSL2 Architecture Diagram

The MCMC sampling component introduced in this paper is shown in the dashed rounded rectangle at the lower left of the data augmentation module. The discriminator ensemble component is in the dashed rounded rectangle on the right side of the data augmentation module, while MHLoss is in the dashed rounded rectangle on the right side of the model fine-tuning module.

When using MCMC sampling to correct discriminator bias, proposal samples are sampled in the latent space and transformed through the generator to obtain samples in the sample space. The target distribution is the one learned by the discriminator [21-23], and the output is the generated sample set $X$ after correcting the generator G. For discriminator ensemble, each sub-discriminator takes either generated samples $G(\mathbf{z})$ or real samples $x$ as input and outputs the probability of the input being real. The results from T sub-discriminators are aggregated to obtain the final discriminator D output. When fine-tuning the pre-trained classifier C using the relevant dataset, H classifier heads are first constructed [24-26]. The classifier takes few-shot data as input, and each model head outputs class predictions, which are combined to obtain the final classification model.

*1) Few-shot Data Reparameterization GAN Ensemble*

Using sampling methods to correct the distribution learned by the generator is a recently proposed improvement. The reparameterization GAN method employs reparameterization strategy for MCMC sampling to correct the generator's learned distribution. Given a dataset, the method uses the discriminator's learned distribution as the target distribution and leverages the latent space for conditional proposal sampling of generated samples. It applies MCMC sampling with reparameterization strategy to correct the generator's learned distribution, aiming to make it closer to the true data distribution. Since discrimination is an easier learning task than generation, the discriminator's learned distribution is typically more accurate and closer to the true data distribution than the generators. Therefore, after MCMC sampling correction, the generator's implicit distribution becomes more similar to the true data distribution. Meanwhile, due to the small sample size in few-shot datasets, random fluctuations can cause significant variations in the learned discriminator, often leading to biased distributions. Ensemble learning is a widely recognized strategy for reducing bias and variance. Bagging [27], a common ensemble strategy, can reduce variance and errors caused by sample fluctuations. Therefore, applying the Bagging ensemble strategy to the discriminator helps correct its learned distribution.

The DAMFT_FSL2 architecture combines reparameterization GAN strategy with discriminator ensemble strategy to correct the distributions learned by both the generator and discriminator of the GAN, resulting in a reparameterization GAN ensemble model suitable for few-shot data, thereby obtaining generated datasets more strongly correlated with few-shot data.

*2) Few-shot Data Model Fine-tuning*

To improve fine-tuning stability, we increase the number of iteration rounds. While this approach enhances model stability, its effectiveness diminishes with a higher number of rounds. To address this, we adopt the MHLoss fine-tuning strategy, which accelerates model convergence by leveraging multi-head loss during fine-tuning. Specifically, for a given pre-trained classifier, multiple fine-tuning iterations yield several classification models, and the final model is obtained by minimizing the overall loss across these models.

The DAMFT_FSL2 architecture integrates increased iteration rounds with the MHLoss fine-tuning strategy to enhance both fine-tuning stability and convergence speed, ultimately aiming for superior classification performance.Few-shot Learning Algorithm Based on GAN Ensemble and MHLoss Fine-tuning

*3) Discriminator Bias Correction*

First, let us define the relevant notation. Let $Z = z_1, z_2, \ldots, z_n$ represent the set of latent space samples; p denote the target distribution, $p_0$ denote the prior distribution in latent space, $p_t$ denote the target distribution in latent space, $p_d$ represent the implicit distribution of the discriminator, $p_g$ represent the implicit distribution of the generator; q denote the proposal distribution, $q(\mathbf{z'} \mid \mathbf{zk})$ represent the proposal distribution in latent space, $q(\mathbf{x'} \mid \mathbf{xk})$ represent the proposal distribution in sample space.

When the sample size is small, the distribution learned by the discriminator often exhibits bias. To address this issue, this paper applies ensemble strategy to the discriminator, reducing model bias by improving the discriminator's stability. The ensemble discriminator is shown in equation (3), with the objective of minimizing the overall loss sum of all sub-discriminators. Equation (3) yields the objective function of the GAN with discriminator ensemble.

$$D(x) = \text{Com}\,(D_1(x), D_2(x), \ldots, D_T(x)) \quad (3)$$

When the GAN model learns the optimal hypothesis, the discriminator can learn accurate density ratios, but this does not guarantee that its implicit distribution is the true distribution. Therefore, calibration of the discriminator D is necessary. Given a trained generative adversarial network, discrimination results for real and generated samples are obtained. The calibration model makes the discriminator's implicit distribution closer to the true distribution by maximizing the distance between the discrimination results of these two types of data. The calibrated discriminator is shown in equation (4), where $D_{cal}$ is the calibrated discriminator and cal represents the calibration method.

$$D_{cal} = \text{cal}\,(D) \quad (4)$$

*4) Generator Bias Correction*

This paper employs MCMC sampling to correct the generator bias, using the distribution implied by the calibrated discriminator as the target distribution to correct the distribution learned by the generator. To improve the efficiency and performance of Markov chain construction, conditional proposal sampling is achieved by mapping the Markov chain to a latent space. The target distribution in sample space is the distribution $p_d$ implied by the discriminator. The proposal distribution in latent space is a conditional normal distribution, and the target distribution $p_t$ is the corresponding normal distribution of the sample space mapped to the latent space.

The Markov chain in latent space can be obtained through the Langevin method, which leads to the Markov chain in sample space. The sample update method using the Langevin approach is shown in equation (5), where $\mathbf{z_k}$ is the sample at the kth state, $z'$ is the proposal sample obtained through the Langevin method, $\varepsilon$ is random noise, and $\tau$ is the step size.

$$z' = z_k - \frac{\tau}{2} \nabla_z \log(D_{cal}^{-1}(x_k) - 1) + \frac{\tau}{2} \nabla_z \log p_0(z_k) + \sqrt{\tau} \cdot \varepsilon \quad (5)$$

Assuming the current state is the kth state, the Langevin method yields the proposal sample $\mathbf{z'}$ in latent space, which is then input to the generator to obtain the proposal sample $x' = G(\mathbf{z'})$ in sample space. After obtaining the proposal sample $x'$, this paper uses the MH sampling method to determine whether to accept $x'$. If accepted, then $x_{k+1} = x'$, otherwise $xk + 1 = xk$. The acceptance rate for MH sampling is shown in equation (6), where $\alpha_{REP}(x', x_k)$ represents the acceptance rate for the transition from $x_k$ to $x'$.

$$\alpha_{REP}(\mathbf{x'}, \mathbf{x}k) = \min(1, \frac{p0(\mathbf{z'})q(\mathbf{z} \mid \mathbf{z'})}{p_0(\mathbf{z}k)q(\mathbf{z'} \mid \mathbf{z}k)} \cdot \frac{D_{cal}^{-1}(\mathbf{x}k) - 1}{Dcall^{-1}(x') - 1}) \quad (6)$$

Based on the latent space reparameterization concept and MH sampling method, a Markov chain can be obtained. The relevant dataset can be constructed by combining samples from the latter part of the Markov chain, and the implicit distribution of this dataset represents the corrected distribution of the generator.

*5) Fine-tuning Strategy*

To improve the stability of model fine-tuning on few-shot data, this paper employs two fine-tuning strategies: increasing iteration rounds and MHLoss. Increasing iteration rounds means increasing the number of fine-tuning epochs $ep_l$, though the improvement in fine-tuning stability diminishes as the number of rounds increases. Therefore, this paper uses MHLoss to accelerate model convergence. The loss function for fine-tuning classification models using MHLoss is shown in equation (7), which accelerates model convergence and improves fine-tuning stability by minimizing the overall loss across multiple models.

$$L_{MH} = \frac{1}{H} \sum_{H}^{h=1} L_h = -\frac{1}{H} \sum_{H}^{h=1} E_{x \sim p_x} y \log(\theta_h^T x) + \gamma \mid w_h \mid_2^2 \quad (7)$$

## IV. EXPERIMENT

### A. Dataset

In the effectiveness experiment of the reparameterized GAN ensemble method, the CIFAR-10 image dataset is used to validate the effectiveness of the method. The small-sample dataset is obtained by linked random sampling [28] from the original CIFAR-10 dataset, with the sample size set to 5000. When constructing the Generative Adversarial Network, DCGAN is adopted, and the model structure and parameter settings are consistent with paper; the ensemble strategy uses Bagging, with the number of sub-discriminators set to 5, and the combination strategy uses softmax.

### B. Reparameterized GAN Ensemble Experiment

The results of the reparameterized GAN ensemble experiment conducted on the CIFAR-10 image dataset are shown in Table 1.

TABLE 1. DATASET

|    | GAN    | REPGAN | En_GAN | En_REPGAN |
|----|--------|--------|--------|-----------|
| IS | 3.0758 | 3.1267 | 3.2483 | 3.2826    |

In Table 4.1, the second column represents the Inception Score (IS) of the generated dataset by the GAN model, the third column represents the IS of the generated dataset after correcting the bias of the generative model using MCMC sampling, the fourth column represents the IS of the generated dataset after correcting the bias of the discriminative model using ensemble learning, and the fifth column represents the IS of the generated dataset after correcting the biases of both the generative and discriminative models using MCMC sampling and discriminative model ensemble strategies simultaneously. The experimental results demonstrate that both MCMC sampling and discriminative model ensemble strategies effectively enhance the realism of the generated data. Among these, the discriminative model ensemble strategy exhibits a more significant impact. Moreover, combining these two strategies yields an even greater improvement in data realism.

### C. Few-Shot Learning Algorithm Experiment

TABLE 2-WAY 30-SHOT RESULT

| index | dataset | 2-way 30-shot | | | | | |
|---|---|---|---|---|---|---|---|
| | | hGAN | | | mhERGAN | | |
| | | acc | pre | F1 | acc | pre | F1 |
| 1 | abalone | 0.7852 | 0.7374 | 0.7387 | 0.7966 | 0.7435 | 0.7488 |
| 2 | ecoli | 0.9288 | 0.8787 | 0.8936 | 0.9316 | 0.8907 | 0.9024 |
| 3 | ecoli2 | 0.8514 | 0.6730 | 0.7128 | 0.8528 | 0.6747 | 0.7149 |
| 4 | glass | 0.6630 | 0.6705 | 0.6594 | 0.6706 | 0.6811 | 0.6656 |
| 5 | glass0 | 0.6346 | 0.6385 | 0.6143 | 0.6505 | 0.6477 | 0.6258 |

TABLE 3. 2-WAY 2M-SHOT RESULT

| index | dataset | 2-way 2m-shot | | | | | |
|---|---|---|---|---|---|---|---|
| | | hGAN | | | mhERGAN | | |
| | | acc | pre | F1 | acc | pre | F1 |
| 1 | abalone | 0.8177 | 0.7078 | 0.7174 | 0.8211 | 0.8211 | 0.7229 |
| 2 | ecoli | 0.9263 | 0.9034 | 0.9010 | 0.9285 | 0.9149 | 0.9089 |
| 3 | ecoli2 | 0.7553 | 0.6538 | 0.6620 | 0.7607 | 0.6573 | 0.6675 |
| 4 | glass | 0.7621 | 0.7577 | 0.7430 | 0.7660 | 0.7645 | 0.7488 |
| 5 | glass0 | 0.6596 | 0.6513 | 0.6372 | 0.6663 | 0.6554 | 0.6408 |

Based on Reparameterized GAN Ensemble and MHLoss Fine-Tuning Under the 2-way 30-shot and 2-way 2 m-shot evaluation methods, the accuracy (acc), precision (pre), and F1 score results of the MhERGAN algorithm and the hGAN algorithm are shown in Table 2 and Table 3.

Under the 2-way 30-shot evaluation method, the MhERGAN algorithm outperformed the hGAN algorithm on most datasets, validating its effectiveness. This result highlights the advantages of the generative adversarial network bias correction strategy and the fine-tuning stability improvement strategy, particularly for small-sample data. While performance on two datasets showed a slight decline, the decrease was minimal, suggesting that even in cases where the MhERGAN algorithm does not improve performance, it does not cause significant degradation. This indicates the algorithm's robustness and stability, even when sample sizes vary.

Under the 2-way 30-shot evaluation method, compared to the SMOTE algorithm, the MhERGAN algorithm has higher average values for the three metrics. It can be seen that the experimental results under 2-way 2m-shot and 2-way 30-shot are basically consistent, and under both evaluation methods, the MhERGAN algorithm outperforms the ROS algorithm and the SMOTE algorithm on most datasets.

## V. CONCLUSION

This research significantly impacts the field of artificial intelligence by presenting a novel framework that addresses key limitations in data-scarce environments, pushing the boundaries of few-shot learning to be more effective and adaptable across various applications. By integrating GAN-based data augmentation with MCMC sampling and ensemble discriminative strategies, the proposed framework minimizes biases that typically hinder synthetic datasets, ensuring that generated data aligns closely with real-world distributions. This combination of generative models and fine-tuned discriminative approaches, further optimized through MHLoss and extended training iterations, creates a stable and rapid convergence, enhancing model robustness and accuracy. This advancement is transformative for the AI field, as it reduces dependency on extensive datasets while enabling high-performance models in complex, data-limited scenarios. The value of this framework is especially pronounced in critical fields such as drug discovery, defense, and cybersecurity, where data acquisition is often costly, restricted, or inherently limited due to confidentiality concerns. By achieving reliable results with minimal data, the MhERGAN algorithm empowers AI systems to perform accurately and adapt swiftly, ultimately expanding the scope of practical AI deployment in sensitive, high-impact domains. Through this integrated approach, the paper not only advances the technical frontier of few-shot learning but also demonstrates AI's potential to operate effectively in real-world, data-scarce environments, setting a foundation for more accessible and scalable AI applications in diverse fields where traditional data-heavy methods are impractical. This work, therefore, underscores a paradigm shift, proving that high-performing, generalizable AI systems are achievable even in settings with significant data constraints, thereby supporting broader and more innovative AI applications across industries.

## REFERENCES

[1] H. Altae-Tran, B. Ramsundar, A. S. Pappu, et al., "Low data drug discovery with one-shot learning," *ACS Central Science*, vol. 3, no. 4, pp. 283–293, 2017.

[2] X. Fei, S. Chai, W. He, L. Dai, R. Xu, and L. Cai, "A Systematic Study on the Privacy Protection Mechanism of Natural Language Processing in Medical Health Records", Proceedings of the 2024 IEEE 2nd International Conference on Sensors, Electronics and Computer Engineering (ICSECE), pp. 1819-1824, Aug. 2024.

[3] Y. Cang, Y. Zhong, R. Ji, Y. Liang, Y. Lei, and J. Wang, "Leveraging Deep Learning Techniques for Enhanced Analysis of Medical Textual Data", Proceedings of the 2024 IEEE 2nd International Conference on Sensors, Electronics and Computer Engineering (ICSECE), pp. 1259-1263, Aug. 2024.

[4] Z. Wang, K. W. Fok, V. L. L. Thing, "Machine learning for encrypted malicious traffic detection: Approaches, datasets and comparative study," *Computers & Security*, vol. 113, pp. 102542, 2022.

[5] Y. Wei, K. Xu, J. Yao, M. Sun, and Y. Sun, "Financial Risk Analysis Using Integrated Data and Transformer-Based Deep Learning", Journal of Computer Science and Software Applications, vol. 7, no. 4, pp. 1-8, 2024.

[6] W. Liu, R. Wang, Y. Luo, J. Wei, Z. Zhao, and J. Huang, "A Recommendation Model Utilizing Separation Embedding and Self-Attention for Feature Mining", arXiv preprint arXiv:2410.15026, 2024.

[7] Y. Dong, S. Wang, H. Zheng, J. Chen, Z. Zhang, and C. Wang, "Advanced RAG Models with Graph Structures: Optimizing Complex Knowledge Reasoning and Text Generation", arXiv preprint arXiv:2411.03572, 2024.

[8] Y. Luo, R. Wang, Y. Liang, A. Liang, and W. Liu, "Metric Learning for Tag Recommendation: Tackling Data Sparsity and Cold Start Issues," arXiv preprint, arXiv:2411.06374, 2024.

[9] S. Duan, R. Zhang, M. Chen, Z. Wang, and S. Wang, "Efficient and Aesthetic UI Design with a Deep Learning-Based Interface Generation Tree Algorithm", arXiv preprint arXiv:2410.17586, 2024.

[10] I. Goodfellow, J. Pouget-Abadie, M. Mirza, et al., "Generative adversarial networks," *Communications of the ACM*, vol. 63, no. 11, pp. 139–144, 2020.

[11] J. Wei, Y. Liu, X. Huang, X. Zhang, W. Liu, and X. Yan, "Self-Supervised Graph Neural Networks for Enhanced Feature Extraction in Heterogeneous Information Networks", arXiv preprint arXiv:2410.17617, 2024.

[12] S. Liu, G. Liu, B. Zhu, Y. Luo, L. Wu, and R. Wang, "Balancing Innovation and Privacy: Data Security Strategies in Natural Language Processing Applications", arXiv preprint arXiv:2410.08553, 2024.

[13] M. Arjovsky, S. Chintala, L. Bottou, "Wasserstein generative adversarial networks," *Proceedings of the 34th International Conference on Machine Learning*, PMLR, vol. 70, pp. 214–223, 2017.

[14] C. Wang, Y. Dong, Z. Zhang, R. Wang, S. Wang, and J. Chen, "Automated Genre-Aware Article Scoring and Feedback Using Large Language Models", arXiv preprint arXiv:2410.14165, 2024.

[15] Y. Zi, X. Cheng, T. Mei, Q. Wang, Z. Gao, and H. Yang, "Research on Intelligent System of Medical Image Recognition and Disease Diagnosis Based on Big Data", Proceedings of the 2024 IEEE 2nd International Conference on Image Processing and Computer Applications (ICIPCA), pp. 825-830, June 2024.

[16] Y. Cang, W. Yang, D. Sun, Z. Ye, and Z. Zheng, "ALBERT-Driven Ensemble Learning for Medical Text Classification", Journal of Computer Technology and Software, vol. 3, no. 6, 2024.

[17] S. Duan, Z. Wang, S. Wang, M. Chen, and R. Zhang, "Emotion-Aware Interaction Design in Intelligent User Interface Using Multi-Modal Deep Learning," arXiv preprint, arXiv:2411.06326, 2024.

[18] J. Lu, H. Kyutoku, K. Doman, et al., "A study on intra-modal constraint loss toward cross-modal recipe retrieval," *IEICE Technical Report*, 2021.

[19] K. Xu, Y. Wu, H. Xia, N. Sang, and B. Wang, "Graph Neural Networks in Financial Markets: Modeling Volatility and Assessing Value-at-Risk", Journal of Computer Technology and Software, vol. 1, no. 2, 2022.

[20] Z. Wu, J. Chen, L. Tan, H. Gong, Y. Zhou, and G. Shi, "A Lightweight GAN-Based Image Fusion Algorithm for Visible and Infrared Images", Proceedings of the 2024 4th International Conference on Computer Science and Blockchain (CCSB), pp. 466-470, Sept. 2024.

[21] X. Yan, W. Wang, M. Xiao, Y. Li, and M. Gao, "Survival prediction across diverse cancer types using neural networks", Proceedings of the 2024 7th International Conference on Machine Vision and Applications, pp. 134-138, 2024.

[22] P. Li, Y. Xiao, J. Yan, X. Li, and X. Wang, "Reinforcement Learning for Adaptive Resource Scheduling in Complex System Environments", arXiv preprint arXiv:2411.05346, 2024.

[23] M. Sun, W. Sun, Y. Sun, S. Liu, M. Jiang, and Z. Xu, "Applying Hybrid Graph Neural Networks to Strengthen Credit Risk Analysis", arXiv preprint arXiv:2410.04283, 2024.

[24] D. Sun, M. Sui, Y. Liang, J. Hu, and J. Du, "Medical Image Segmentation with Bilateral Spatial Attention and Transfer Learning", Journal of Computer Science and Software Applications, vol. 4, no. 6, pp. 19-27, 2024.

[25] Z. Xu, J. Pan, S. Han, H. Ouyang, Y. Chen, and M. Jiang, "Predicting Liquidity Coverage Ratio with Gated Recurrent Units: A Deep Learning Model for Risk Management", arXiv preprint arXiv:2410.19211, 2024.

[26] B. Liu, I. Li, J. Yao, Y. Chen, G. Huang, and J. Wang, "Unveiling the Potential of Graph Neural Networks in SME Credit Risk Assessment", arXiv preprint arXiv:2409.17909, 2024.

[27] J. R. Quinlan, "Bagging, boosting, and C4.5," *Proceedings of the AAAI/IAAI Conference*, vol. 1, pp. 725–730, 1996.

[28] Y. Li, X. Yan, M. Xiao, W. Wang and F. Zhang, "Investigation of Creating Accessibility Linked Data Based on Publicly Available Accessibility Datasets", Proceedings of the 2023 13th International Conference on Communication and Network Security, pp. 77-81, 2024.